# Vision-based Pedestrian Alert Safety System (PASS) for Signalized Intersections

**Mhafuzul Islam***
Ph.D. Student
Center for Connected Multimodal Mobility ($C^2M^2$)
Glenn Department of Civil Engineering, Clemson University
351 Flour Daniel, Clemson, SC 29634
Tel: (864) 986-5446; Email: mdmhafi@clemson.edu

**Mizanur Rahman, Ph.D.**
Postdoctoral Research Fellow
Center for Connected Multimodal Mobility ($C^2M^2$)
Glenn Department of Civil Engineering, Clemson University
127 Lowry Hall, Clemson, SC 29634
Tel: (864) 650-2926; Email: mdr@clemson.edu

**Mashrur Chowdhury, Ph.D., P.E., F.ASCE**
Eugene Douglas Mays Endowed Professor of Transportation
Center for Connected Multimodal Mobility ($C^2M^2$)
Glenn Department of Civil Engineering, Clemson University
216 Lowry Hall, Clemson, SC 29631
Tel: (864) 656-3313; Email: mac@clemson.edu

**Gurcan Comert, Ph.D.**
Department of Computer Science, Physics, and Engineering
Benedict College, 1600 Harden St., Columbia, SC 29206
Tel: (803) 705 4803; Email: gurcan.comert@benedict.edu

**Eshaa Deepak Sood**
M.S. Student
School of Computing, Clemson University
351 Flour Daniel, Clemson, SC 29634
Tel: (864) 643-7284; Email: esood@g.clemson.edu

**Amy Apon, Ph.D.**
Professor and Director
School of Computing, Clemson University
100 McAdams Hall, Clemson, SC 29634
Tel: (864) 656-5769; Email: aapon@clemson.edu

*Corresponding Author

**ABSTRACT**

Although Vehicle-to-Pedestrian (V2P) communication can significantly improve pedestrian safety at a signalized intersection, this safety is hindered as pedestrians often do not carry hand-held devices (e.g., Dedicated short-range communication (DSRC) and 5G enabled cell phone) to communicate with connected vehicles nearby. To overcome this limitation, in this study, traffic cameras at a signalized intersection were used to accurately detect and locate pedestrians via a vision-based deep learning technique to generate safety alerts in real-time about possible conflicts between vehicles and pedestrians. The contribution of this paper lies in the development of a system using a vision-based deep learning model that is able to generate personal safety messages (PSMs) in real-time (every 100 milliseconds). We develop a pedestrian alert safety system (PASS) to generate a safety alert of an imminent pedestrian-vehicle crash using generated PSMs to improve pedestrian safety at a signalized intersection. Our approach estimates the location and velocity of a pedestrian more accurately than existing DSRC-enabled pedestrian hand-held devices. A connected vehicle application, the Pedestrian in Signalized Crosswalk Warning (PSCW), was developed to evaluate the vision-based PASS. Numerical analyses show that our vision-based PASS is able to satisfy the accuracy and latency requirements of pedestrian safety applications in a connected vehicle environment.

## 1. INTRODUCTION

Road traffic injuries are a leading cause of death worldwide, accounting for more than 1.35 million deaths annually. Approximately 54% of these deaths involve pedestrians, cyclists, motorcyclists, and road workers, a cohort that is collectively known as Vulnerable Road Users (VRUs) *(WHO, 2018)*. In the United States, traffic-related fatalities are the leading cause of death for people ages one to forty-four, thus, adding to the category of "unintended injury", which is the third highest cause of death overall *(NTHSA, 2017; NTHSA, 2018)*. An average of 69,000 pedestrians is injured annually on U.S. roadways, where pedestrians account for 14% of U.S. road fatalities *(NTHSA, 2018)*. Since 2009, the number of pedestrian fatalities has been rising, reaching over 6,000 in 2017, an increase of over 30%. This cause of death, which was 12% of total traffic-related fatalities in 2009, now represents over 16% of total traffic deaths as of 2017 (*NTHSA, 2017)*.

Safety research for VRUs in a connected vehicle environment is one of the emerging research topics given that majority of the VRUs-involved incidents can be prevented through early warnings to both vehicles and VRUs *(Rahman et al., 2018; Sewalkar and Seitz, 2019; Hoye and Laureshyn, 2019)*. Recent research shows that enabling Dedicated Short-Range Communication (DSRC), a low latency communication medium for safety applications, in a pedestrian hand-held device (e.g., smartphones) can increase pedestrian safety significantly through vehicle-to-pedestrian (V2P) communication *(*Wu et al., *2014)*. The use of 5G, a nascent low latency emerging communication medium, is also useful in safety-critical connected vehicle applications. The DSRC/5G-enabled V2P communication gives a 360º view in which both the driver and the pedestrian are warned of possible collision risks using DSRC-based personal safety messages (PSMs) *(Sewalkar and Seitz, 2019)*. However, all pedestrians are unlikely to carry a DSRC-enabled device with an activated pedestrian safety application. In addition, the high data exchange latency of current cellular communication networks, such as LTE, makes them ill-suited for pedestrian safety applications *(Dey et al., 2014;* Wu et al., *2014, Xu et al., 2017)*.

To overcome the aforementioned limitations and increase pedestrian safety, video data from traffic cameras at a signalized intersection are used to detect and generate personal safety messages (PSMs). PSMs are then used to provide safety alerts about the pedestrians to approaching nearby connected vehicles. Using the fundamental pedestrian information (e.g., velocity and location), the DSRC-enabled roadside infrastructure generates and broadcasts PSMs to the approaching connected vehicles. Any pedestrian-related safety applications can use these PSMs and warn approaching connected vehicles at a safety-critical roadway section or at an intersection. The approach requires no DSRC/5G-enabled hand-held devices to be carried by pedestrians.

In this study, we develop a vision-based pedestrian alert safety system (PASS) that can generate personal safety messages (PSMs) and pedestrian safety alerts using PSMs in real-time (every 100 milliseconds) to improve pedestrian safety at an intersection. Our approach generates PSMs following the Society of Automotive Engineers (SAE) J2945 standard *(SAE 2018)*. We evaluate the accuracy of generated PSMs by comparing with the field collected ground truth data and the data from the existing DSRC based hand-held device. Furthermore, we conducted a real-world experiment to evaluate the efficacy of the PASS by developing a pedestrian safety application, Pedestrian in Signalized Crosswalk Warning (PSCW) *(ARC-IT, 2019)*.

The remainder of the paper is structured as follows. In Section 2, the existing research on pedestrian detection methods, the pedestrian safety-related studies in a connected vehicle environment, and standards for PSMs are described. In Section 3, the method in developing for vision-based deep learning technique is detailed, which is used to generate PSMs and pedestrian safety alert to improve pedestrian safety at a signalized intersection. In Section 4, the evaluation of the generated PSMs are presented, including all evaluation scenarios, data description, real-world experimental descriptions, and results. Section 5 discusses the evaluation of PASS using Pedestrian at Signalized Crosswalk Warning (PSCW) application. Finally, the conclusions from the study and possible future research are discussed in Section 6.

## 2. RELATED STUDIES

The advent of connected vehicle technologies and deep learning enables the improvement of pedestrian safety. In this section, we explore the existing research on pedestrian detection methods pedestrian safety-related studies in a connected vehicle environment, and standards for PSMs.

### 2.1 Pedestrian detection methods

Many studies involving the use of different sensors, such as ultrasonic sensors, radar, laser, and cameras have been undertaken to develop pedestrian detection schemes at intersections to improve safety *(Bu et al., 2005; Alonso et al., 2007; Ismail et al., 2009; Boudet and Midenet, 2009; Ismail et al., 2010; Dollar et al., 2012; Benson et al., 2014; Alahi et al., 2014; Fang et al., 20014; Xu et al., 2017*). Although ultrasonic sensors based on sound waves can detect pedestrians up to 30 feet, the sensor set-up angle, pedestrian clothing, and weather conditions affect the performance *(Beckwith et al., 1998)*. A Doppler ultrasonic sensor detects an object based on the relative velocity between the object and the sensor while newer ultra-wideband (UWB) technology can sense pedestrians within an accuracy of one inch. In addition, radar detectors operate reliably under inclement weather and environmental conditions. Infrared pulse laser scanners are accurate and informative in terms of detection accuracy of a pedestrian; however, they do require high computational power and the accuracy drops in foggy and snowy conditions. Conversely, the computer-vision based technique is hindered by high false positives, high miss rates, the difficulty in detecting stationary pedestrians and in detecting patterns of moving pedestrians (walking, jogging, jumping), detection of body parts, occlusion level, and computational complexity of processing for detecting pedestrians *(Gandhi and Trivedi, 2006; Ismail et al., 2009)*. In their comparison of thermal infrared radar and computer vision-based pedestrian detection approaches in terms of capabilities, cost, accuracy, and computational complexity levels Gandhi and Trivedi (2006) found that a combination of different technologies increased detection performance.

Recently, researchers have begun to utilize deep learning methods for pedestrian detection *(Zhao and Thorpe, 2000; Wang et al., 2012; Zeng et al., 2013; Zeng et al., 2014; Zhang et al., 2016)*. Using a deep learning classifier for pedestrian detection, Zeng et al. *(2014)* developed an adaptive method that is transferrable to different locations without retraining. The method was also deemed superior to existing counterparts. Also, their Multi-layer Convolutional Neural Network (CNN) method did not assume the same data characteristics for training and testing datasets and aimed to determine the characteristic of the testing data in a rolling window approach. Presented methods were trained on the INRIA dataset and evaluated using the MIT and CUHK datasets (*Zeng et al., 2014*). Similarly, in a real-time implementation study of related models, Angelova et al. presented a Deep Network Cascades for both accurate and fast pedestrian detection *(*Angelova et al., *2015)*. They noted that the methods used in Dollar et al. *(2012)* and Benenson et al. *(2014)* could be implemented in a real-time pedestrian detection environment with high miss rates (MR). Their method required a computational cost of 0.067 s (15 frames per second (fps)) and 26.2% MR. Although the hybrid Katamari method gave 22% MR, its computational cost was not listed. Except for the WordChannels method with an MR of 42.3%, it is also possible to run other methods over one second to process a frame. Nevertheless, the range of detection capabilities was not shown in Benenson et al., 2014. More recently, Cai et al. 2015 developed a compact-deep model with which they detected pedestrians within one second at approximately 10% MR *(Cai et al., 2015; Cai et al., 2016)*. The advent of higher processing GPU-enabled devices has improved pedestrian detection in terms of both accuracy and computation time. Indeed, the following state-of-the-art deep learning based object detection frameworks are now regularly used for real-time performance: Region-Convolutional Neural Network (R-CNN) *(Girshick et al., 2014*), Fast R-CNN *(Girshick R., 2015),* Faster R-CNN *(Hanna and Cardillo, 2013),* Single Shot MultiBox Detector (SSD) *(Liu et al., 2016),* and You Only Look Once - Version 3 (YOLOv3) *(Redmon et al., 2018)*. Considering the accuracy of pedestrian detection, the YOLOv3 shows superior performance with 81% accuracy and prediction detection time of 51 milliseconds (ms) (approximately 20 fps), which exceeds all other state-of-the-art pedestrian detection deep learning models *(Redmon et al., 2018)*. As safety-critical applications require a high detection accuracy, we developed a strategy to achieve a high detection accuracy using a deep learning method that also satisfies real-time latency requirements.

### 2.2 Pedestrian safety-related studies for the connected vehicle environment

Safety research for vulnerable road users (VRU) in a connected vehicle environment using vehicle-to-pedestrian (V2P) communication, is one of the emerging topics given the fact it is possible to prevent most incidents involving VRUs through giving early warnings to both vehicles and VRUs. For example, investigations have been undertaken to use V2P communications via DSRC, Wi-Fi, and cellular communication technologies to ensure pedestrian safety *(Wu and Nevatia, 2007; Wu et al., 2013; Anaya et al., 2014; Bagheri et al., 2014)*. Such DSRC communications can warn both users within one second, given that the environment is under an ideal setup *(Wu and Nevatia, 2007)*. Compared to existing WiFi and Cellular LTE technologies, the communication range of DSRC-enabled devices is 300 meters, which makes it a much more rapid two-way communication option for information-sharing *(Sewalkar and Seitz, 2019)*. However, the pedestrian's position accuracy depends on the Global Navigation Satellite System (GNSS) and Global Positioning System (GPS) accuracies, low or no message packet drops, and low latency. Although lateral GPS accuracy is sufficient, lane level accuracy is not usually met. The efficacy of Wi-Fi in satisfying pedestrian safety applications within a certain distance, coverage, latency, and drop rates was discussed in previous studies. Anaya et al. *(2014)* found the GPS accuracy at 10 m, clearly inadequate for pedestrian safety applications. In another study, Tahmasabi-Sarvestani et al. *(2017)* presented an active safety system for VRUs using DSRC-enabled smartphone and DSRC-equipped vehicle that consist of sensors to observe the surrounding environment of the vehicle. To reduce energy consumption, they also adopted an adaptive message sending strategy for different locations at varying time intervals. The reported method lowered the probability of hazardous situations considering situational awareness. However, to the best of our knowledge, in the absence of a pedestrian hand-held device in a connected vehicle environment, the improvement of pedestrian safety is not studied before. In this study, we develop a framework to improve pedestrian safety using a vision-based deep learning approach where it is assumed that pedestrians do not carry a DSRC-enabled hand-held device.

### 2.3 The standard for Personal Safety Messages (PSMs)

A standard message set for personal safety messages (PSMs) makes the data communication interoperable between the DSRC-enabled pedestrian hand-held devices and the connected vehicles. PSMs are transmitted by the hand-held device carried by a roadway user, such as pedestrians and bicyclists, both of which are known as Vulnerable Road Users (VRUs). The PSMs for these VRUs are defined by the SAE J2945 standard *(SAE, 2019)* for safety data communication between the VRUs and other connected components (e.g., vehicle and traffic signals). Although SAE J2945 defines the PSM standard, SAE J2735 *(SAE, 2016)* defines the format and structure of the message, data frames, and data elements for exchanging data between VRUs and vehicles and between VRUs and infrastructure. Conversely, SAE J2945 considers each of the data elements defined in SAE J2735. The standard data format and structure of the PSM are detailed below in Table 1.

**Table 1: PSM standard based on SAE J2945 and SAE J2735** *(SAE 2016 and 2019)*

| Data Element | Description |
|---|---|
| **Personal Device User Type** | It is a data element set in accordance with the user type |
| **Message ID** | It is a data element used in each message to define the type of message. It is always the first value inside the message that tells the receiving application how to interpret the remaining bytes. |
| **Timestamp** | It is a single value data element consisting of integer values from zero to 60999 representing the milliseconds within a minute. A leap second is represented by the value range 60001 to 60999. |
| **Message Count** | It is a data element used to provide a sequence number within a stream of messages with the same DSRC message ID and from the same sender. |
| **Temporary ID** | It is a four-byte random device identifier called the temporary id, which is a value for a mobile OBU that periodically changes to ensure overall pedestrian anonymity. |
| **Latitude** | It is the data element used to represent the geographic latitude of the pedestrian expressed in 1/10th integer micro degrees as a 32-bit value concerning the horizontal datum in use. |
| **Longitude** | It is a data element used to represent the geographic longitude of the pedestrian expressed in 1/10th integer micro degrees as a 32-bit value and concerning the horizontal datum in use. |
| **Elevation** | It is a data element used to represent the geographic position of the pedestrian above or below the referenced ellipsoid. The 16-bit number has a resolution of 1 decimeter and represents an asymmetric range of positive and negative values. |
| **Positional Accuracy** | It is a data element consisting of a four-octet field of packed data of various parameters of quality used to model the accuracy of positional determination for each given axis. |
| **Velocity** | It is a data element that represents the velocity of pedestrian expressed in unsigned units of 0.02 meters per second. |
| **Heading** | It is a data element used to represent the current heading direction of the pedestrian expressed in unsigned units of 0.0125 degrees from north. |

## 3. VISION-BASED PEDESTRIAN ALERT SAFETY SYSTEM (PASS)

We develop a system that uses a real-time camera feed, generates PSMs using a vision-based deep learning model, and generates pedestrian safety alerts or pedestrian collision warnings at a signalized intersection. From this point forward, we will use the term “pedestrian” to represent a VRU and vice-versa. As shown in Table 1 of Subsection 2.3, the pedestrian’s latitude, longitude, velocity, and heading direction are the key data elements for generating PSMs as these elements require calculation every one-tenth of a second for each pedestrian. We generate the value of these key data elements based on the SAE J2945 standard for each pedestrian in real-time. DSRC Message-ID, Timestamp, Message Count, and Temporary ID are assigned for every PSM packet sequentially for a specific pedestrian. On the other hand, the elevation is the static value for a particular signalized intersection, and the positional accuracy is determined based on the accuracy of pedestrian location information (i.e., longitude and latitude). As presented in Figure 1, prior to generating the PSMs, we resize a given image to reduce computational time for pedestrian detection. We then develop an approach to achieve high pedestrian detection accuracy via calibration of a pedestrian detection model and removal of duplicate pedestrian detection bounding boxes using non-max suppression technique. We next use an image masking technique to filter the pedestrians using a roadway mask image to reduce unwanted pedestrian detection. We then transform the perspective of an image to localize a pedestrian accurately and finally calculate the location, velocity, and heading information of a pedestrian to construct a PSM. Then, the constructed PSMs are used by pedestrian safety application to generate pedestrian collision warnings. The details of this framework are described in the following subsections.

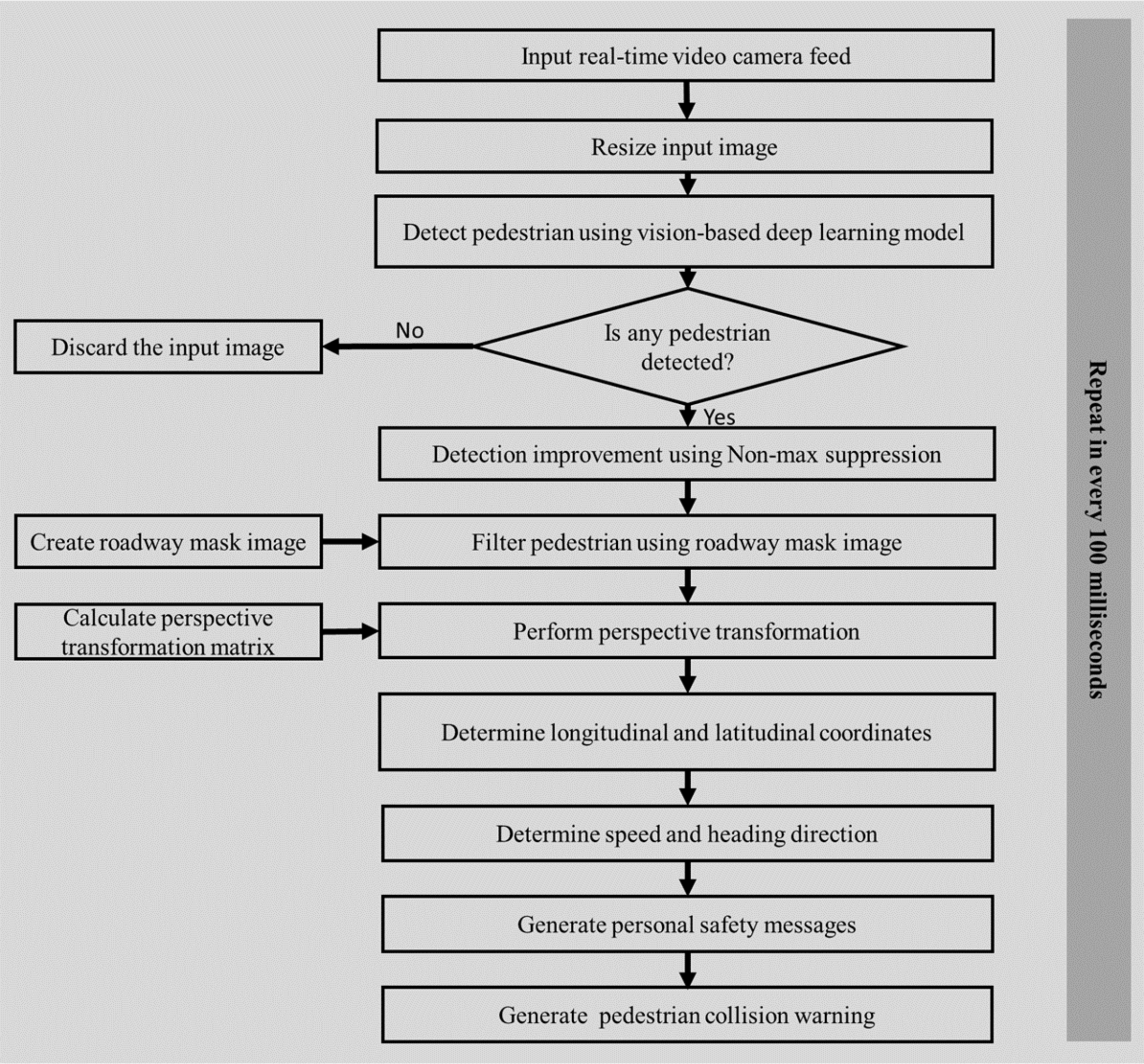

**Figure 1: Flowchart for the Pedestrian Alert Safety System (PASS) using Personal Safety Messages (PSMs)**

### 3.1 Deep Learning Model

A high pedestrian detection accuracy and a low computational time are the key motivations and challenges for implementing a vision-based deep learning model for safety-critical applications.

Based on our literature review, YOLOv3 model can provide the highest detection accuracy (81%) along with a very low computational time (51ms) *(Redmon et al., 2018)*. However, the accuracy of a vision-based pedestrian detection model also varies in different weather conditions (e.g., in cloudy, rainy and snowy conditions) with various numbers of pedestrians. As such, for our purposes, it is necessary to achieve higher pedestrian detection accuracy by retraining the YOLOv3 model with the video data from a signalized intersection. The pedestrian detection model gives an output in a matrix form of a bounding box for each detected pedestrian, with the bounding box defined as a matrix in the form of $(C, P, P_x, P_y, P_h, P_w)$. Here, $C$ is the type of object detected. In our case, if the detected object is a pedestrian, then *C=1*, otherwise *C=0*. If *C = 0*, the remaining attributes are ignored. If *C=1*, then $P$ is the confidence score of the detected pedestrian. $P_x$ is the pixel value on the x-axis of the bottom-center corner of the bounding box and $P_y$ is the pixel value on the y-axis of the bottom center corner of the bounding box of the pedestrian. $P_h$ and $P_w$ are the height and width of the bounding box, respectively. For example, the pictorial representation of the input and output of the YOLOv3 model is shown in Figure 2. The input image size is 640×480 pixels and the output of the YOLOv3 model is $(C, P, P_x, P_y, P_h, P_w) = (1, 0.77, 0.13, 0.47, 0.08, 0.11)$. Here, the values of $(P_x, P_y, P_h, P_w)$ represent the relative position $(P_x, P_y)$ of the bounding box with respect to the image height $(P_h)$ and image width $(P_w)$. Information related to the relative position of the bounding box are important for localizing a pedestrian in terms of latitude and longitude. However, even after retraining any pedestrian detection model, the model predicts duplicate bounding boxes for a single pedestrian. Given that duplicate bounding boxes reduce the pedestrian detection accuracy significantly, a non-max suppression technique is useful for preventing such duplication of bounding boxes.

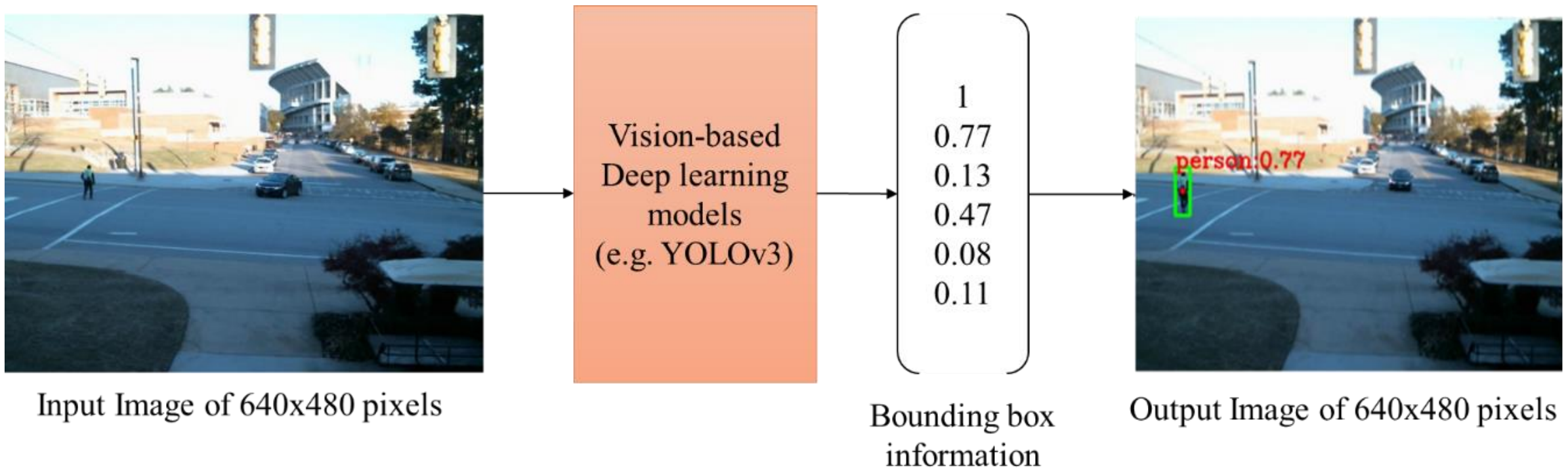

**Figure 2: An example of the YOLOv3 model input and output.**

### 3.2 Non-max Suppression Method

We use a non-max suppression algorithm to remove the duplicate bounding boxes and to improve the accuracy of pedestrian detection. This algorithm prevents overlap of the bounding boxes with a high confidence score and removes the other overlapping regions that are characterized by an Intersection-over-Union (IoU) value bigger than 0.5 *(Rothe et al., 2014)*. The intersection over union is defined by Eq. (1).

$$IoU_{i,j} = \frac{O_{i,j}}{U_{i,j}} \tag{1}$$

where, $IoU_{i,j}$ is the IoU value of bounding box $i$ and $j$; $O_{i,j}$ is the overlapped area of bounding boxes $i$ and $j$ ; and $U_{i,j}$ area of the union of bounding boxes $i$ and $j$. Non-max suppression is computationally inexpensive compared to the sliding windows search method and is an integral part of many deep learning based object detection methods *(Rothe et al., 2014)*.

### 3.3 Image Masking

Although the pedestrian detection models detect any pedestrian within the video camera frame, some pedestrians within that frame may be outside of the boundary of both the roadway and crosswalk. To generate PSMs, we only detect pedestrians within the roadway and crosswalk as they are within the safety-critical region of the intersection. Thus, we use a road mask or binary image of the road and filter the pedestrians outside of the region of the road. A road mask binary image is an image where each pixel is either 0 or 1. If a pixel belongs to the road, then the value of that pixel is 1; otherwise 0. For example, if the pixel location of the pedestrian is $(P_x, P_y)$ and the road mask image pixel value, $I[P_x, P_y] = 1$, the pedestrian is taken into consideration for the next step. A pixel-wise binary image generation tool, such as the online tool LabelBox *(LabelBox, 2019)*, is used to generate this binary mask image.

### 3.4 Perspective Transformation

The accurate location of a pedestrian is required for PSMs to generate collision warnings between pedestrians and vehicles. In addition to creating the binary image, we perform a perspective transformation of each image of a video camera's feed. However, the collected raw images from the roadside camera do not provide the top view of the intersection. Therefore, we incorporate a perspective transformation to create an image, which provides a top view of an intersection. Using this approach, we can select a region of interest of a roadway segment to transform the perspective. The perspective transformation is computed on each pixel of the image using Eq. (2) as follows to generate a top-view of the image:

$$[x', y', z']^T = M \cdot [x, y, z]^T \quad (2)$$

where *x*, *y*, and *z* are the pixel coordinates of an image; *M* is the perspective transform matrix; $x'$, $y'$, and $z'$ are the coordinates of a new location of the pixel after transformation.

### 3.5 Pedestrian Localization

We define four world coordinates of an intersection that corresponds to four corners of an image of the intersection. 1, 2, 3 and 4 represent the top-left, top-right, bottom-right and bottom left corners, respectively, of an image, and $W_1, W_2, W_3, and\, W_4$ are the corresponding world coordinates. Each world coordinate, $W_i$ = $(W_{x,i}, W_{y,i})$ contains the latitude $(W_{x,i})$ and longitude $(W_{y,i})$; where $i$ = [1, 2, 3, and 4]. After performing a perspective transformation, we obtain a 3×3 perspective transformation matrix, *M*. If we multiply this matrix with the pixel coordinate, $(P_x, P_y)$, of the bounding box, which is generated from pedestrian detection models, we can calculate the new transformed pixel-coordinate, $(px, py)$, of a pedestrian using Eq. (3):

$$[px, py, 1]^T = M \cdot [P_x, P_y, 1]^T \quad (3)$$

where, $(px, py)$ is the new transformed pixel-coordinate of the pedestrian. Next, we perform a coordinate transformation using Eq. (4) and Eq. (5) from pixel coordinates to world coordinates (i.e., latitude and longitude) relative to the top-left corner of the image using a linear transformation.

$$L_x = \left(\frac{\Delta Wx}{\Delta px}\right) \times px + W_{x1} \tag{4}$$

$$L_y = \left(\frac{\Delta Wy}{\Delta py}\right) \times py + W_{y1} \tag{5}$$

where, $W_{x1}$ and $W_{y1}$ is the world coordinate of the top left corner of the image representing latitude and longitude, respectively; $\frac{\Delta Wx}{\Delta px}$ represents the rate of change of latitude per pixel of the image; and $\frac{\Delta Wy}{\Delta py}$ represents the rate of change of longitude per pixel of the image. Thus, using these equations, we can calculate the actual world coordinate $(L_x, L_y)$ of a pedestrian.

### 3.6 Pedestrian Velocity and Direction

The walking velocity and heading direction of a pedestrian are two important components of the PSMs. Calculating these criteria requires two consecutive frames from where the location of a pedestrian is known. For example, let $L_1 = (L_{x1}, L_{y1})$ is the location of a pedestrian at time $T_1$, and $L_2 = (L_{x2}, L_{y2})$ is the location of the same pedestrian at time $T_2$. The point-to-point distance between these two locations is calculated using the Haversine formula *(Williams, 2011)* using Eqs. (6-9):

$$a = sin^2\left(\frac{\Delta\varphi}{2}\right) + cos\ \varphi_1 \times cos\ \varphi_2 \times sin^2\left(\frac{\Delta\gamma}{2}\right) \tag{6}$$

$$d = R \times 2atan2\ (\sqrt{a}, \sqrt{1-a}) \tag{7}$$

$$v = \frac{d}{|T2 - T1|} \tag{8}$$

$$h = atan2\ (\Delta\varphi, \Delta\gamma) \tag{9}$$

where, $\varphi_1$ and $\varphi_2$ are the latitude of locations $L_1$ and $L_2$ in degree radians; $\Delta\varphi$ is the difference of latitudes of locations $L_1$ and $L_2$ are in radians; $\Delta\gamma$ is the difference of longitude of $L_1$ and $L_2$ locations in radians; $R$ is the radius of the earth in meters; $d$ is the distance between locations $L_1$ and $L_2$ locations in meters where the $atan2$ function returns a single value $\theta$ such that $-\pi < \theta \leq \pi$ ; $v$ is the pedestrian walking velocity in meters per second (m/s) and $h$ is the pedestrian heading direction in radians.

### 3.7 PSMs Generation

After calculating the location, velocity and heading direction of a pedestrian, we construct the PSMs according to the SAE J2945 standard described in section 2.3. We use the default values while constructing PSMs if the value of that field is unknown or undefined. For example, we use the value of 210 m for elevation attribute as the considered signalized intersection in our study has an elevation of 210 m from sea-level. The positional accuracy was used as 0.54 m from the 1-sigma standard deviation from the collected data. Also, 'Message-ID' is an incremental counter starting from 0 for each message, and 'Temporary ID' is a unique identifier for each pedestrian. The 'Second' attribute is in UTC 13-digits time format to present millisecond-level accuracy. 'Longitude' and 'Latitude' are represented in Global Positioning System (GPS) where 'Velocity' is in meter per second (m/s) format, and for simplicity, 'Heading' is represented in a plain text with any of four of ['East-West', 'West-East', 'North-South', 'South-North']. These attributes are calculated in every one-tenth of a second and used to construct the PSMs, which are then broadcasted to the connected vehicles using a DSRC-enabled device.

### 3.8 Pedestrian Safety Alert Generation

The PSCW application generates pedestrian collision warnings or safety alerts using PSMs from a pedestrian and Basic Safety Messages (BSMs) from a connected vehicle *(USDOT, 2016).* The PSCW application extracts the location and velocity information of the pedestrian and the vehicle from PSMs and BSMs, respectively, to calculate the time-to-collision (TTC) between pedestrian and vehicle (*Karamouzas et al., 2009*). For example, let $L_p = (L_{xp}, L_{yp})$ is the location of a pedestrian, $L_v = (L_{xv}, L_{yv})$ is the location of the vehicle, $V_p = (V_{xp}, V_{yp})$ is the velocity of a pedestrian, and $V_v = (V_{xv}, V_{yv})$ is the velocity of the vehicle. If the location of a pedestrian and a vehicle after time, $t$ in s, are $L_{tp}$ and $L_{tv}$, respectively, then they will colloid each other, if it satisfies the Eq. (10):

$$\left|L_{tp} - L_{tv}\right| = \varepsilon \tag{10}$$

where, $\varepsilon$ is the distance in meters between the center of a vehicle and the pedestrian when they colloid each other. Thus, $\varepsilon = \frac{l}{2}$ and $l$ is the length of a vehicle in meters; the location of a pedestrian after time $t$ is $L_{tp} = L_p + V_p t$, and location of a vehicle after time $t$ is $L_{tv} = L_v + V_v t$. Thus, the Eq. (10) can be written as shown in Eq. (10):

$$\left|(L_p - L_v) + (V_p - V_v)t\right| = \varepsilon \tag{11}$$

If there exist solution of $t$ in s, such that $t \geq 0$, and $t \leq 8$, of the above equation, then this $t$ represents the TTC and we generate a pedestrian safety alert. The maximum value of $t$ = 8 s is taken from the American Association of State Highway and Transportation Officials (AASHTO) guideline for a passenger vehicle's stopping sight distance formula at a signalized intersection (*AASHTO, 2001).* According to the AASHTO guideline, at any roadway design velocity at a signalized intersection, a minimum of 8 s time is desirable for a driver to stop the vehicle without any erratic behavior. The DSRC-enabled roadside device will be used to broadcasts these alerts to connected devices, such as connected vehicles and dynamic message signs. Upon receiving a pedestrian safety alert in the connected vehicle, the vehicle issues an audible warning to the driver regarding a potential vehicle-pedestrian collision.

## 4. EVALUATION OF VISION-BASED PERSONAL SAFETY MESSAGES

The performance of our vision-based PASS depends on the accuracy of the PSMs generated from the video image processing using deep learning based approach. The experimental setup for evaluating PSMs generation are detailed in this section. For our case study, we have selected a T-intersection at Perimeter Road and Avenue of Champions, located at Clemson, South Carolina, U.S., which is a part of Clemson University Connected and Automated Vehicle Testbed (CU-CAVT) *(Chowdhury et. al., 2018*). In the following subsections, we describe the experimental set-up, field data collection and evaluation results of vision-based PSMs following the safety requirements of SAE J2945 standard.

### 4.1 Experimental Set-up

In this experimental setup, we use a video camera that is connected to the roadside data processing device, i.e., Jetson TX2 NVIDIA Pascal™ with 256 NVIDIA CUDA cores and 8GB of GPU memory *(Jetson 2019)*, along with DSRC-enabled roadside unit (RSU) as shown in Figure 3. Because of limited data processing capability of DSRC-enabled RSU, a data processing device is used for running the vision-based PSMs algorithm for detecting pedestrians and constructing PSMs for each pedestrian. After constructing PSMs, the data are transferred to a DSRC-enabled RSU. The Jetson TX2 NVIDIA device and DSRC-enabled RSU are connected using an Ethernet cable. After receiving pedestrian safety alerts from the Jetson TX2 NVIDIA device, the DSRC-enabled RSU broadcasts these pedestrian safety alerts, which are received by the nearby DSRC-enabled connected devices, such as connected vehicles and dynamic message signs. Furthermore, we use a DSRC-enabled hand-held device for pedestrians to collect the pedestrian-related data (e.g., timestamp, longitude, latitude and walking velocity). Later, we use this data for comparing the accuracy of the messages generated from the vision-based PSMs generation approach. The following sections describe the data collection procedure and format of the collected data.

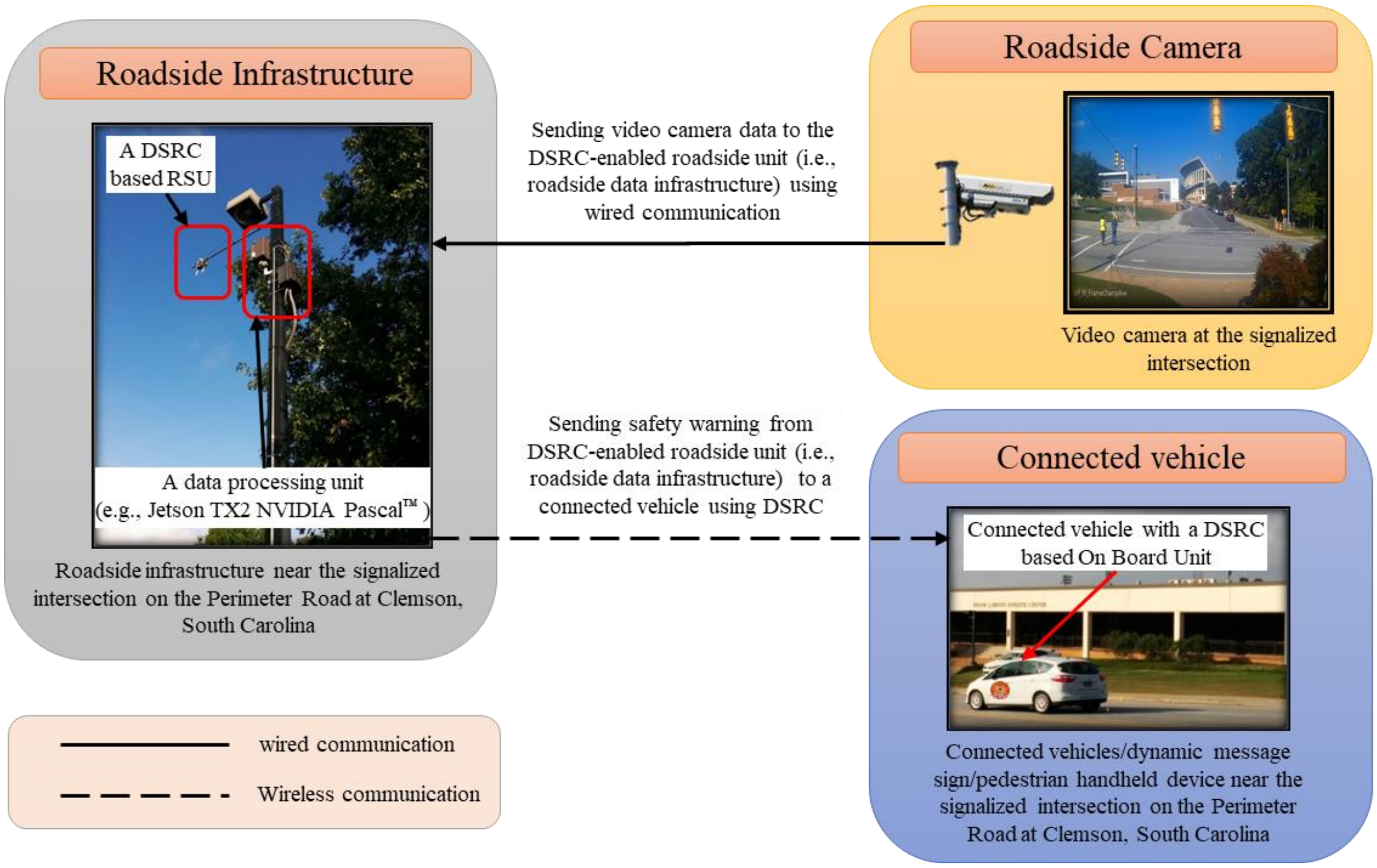

**Figure 3: Experimental setup for generating vision-based pedestrian safety alerts.**

## 4.2 Field Data Collection

Field data is collected for 1) training and testing of the YOLOv3 pedestrian detection model to ensure a high level of pedestrian detection accuracy, which is necessary for safety-critical applications; and 2) the evaluation for generating PSMs, followed by pedestrian collision warnings validation. For training and testing YOLOv3 model, we collected video data related to pedestrian movement from two signalized intersections within the Clemson University Campus: i) Perimeter Road and Avenue of Champions, and ii) College Avenue and SC Highway 93. We selected these two intersections, from which we collected a total of 300 images, to capture a different number of pedestrians and evaluate the performance/accuracy of pedestrian detection. The first intersection was a less busy signalized intersection with the number of pedestrians ranging from one to three at any instance. The second, however, was a busy signalized intersection with the number of pedestrians ranging from three to ten at any instance. The image size of the collected data is 720×576 pixels, which is subsequently reduced to 640×480 pixels to reduce the computational time of pedestrian detection. This reduction is necessary to reduce the computational time for detecting pedestrians for generating PSMs every 100 ms and for running pedestrian safety applications in real-time. Although the video image processing time is proportionate to the size of the image, the pedestrian detection accuracy is reduced with a reduction of image size. Using trial and error, we select the image size of 640×480 pixels, which gives a reasonable trade-off between the image size and the pedestrian detection accuracy.

For the evaluation of generated PSMs, we collected pedestrian-related data from our experiments through the use of 1) a DSRC-enabled hand-held device of pedestrians containing the pedestrian location (i.e., latitude and longitude), velocity, and heading direction; 2) and the generated PSMs using our vision-based deep learning approach. Both the PSMs from DSRC-enabled pedestrian hand-held and our vision-based approach are time-synced using the UTC 13-digits time format *(Schossmaier et al., 1997)*. As shown in Table 2, we collected data for four different pedestrian heading directions: East-West (EW); West-East (WE); North-South (NS); and South-North (SN). To quantify the pedestrian location and velocity estimation accuracy, we collected a total of 225 observations to evaluate both the pedestrian location and walking velocity.

**Table 2: Number of observations for different pedestrian heading direction**

| Pedestrian heading direction | Number of observations |
|---|---|
| **East-West (EW)** | 88 |
| **West-East (WE)** | 59 |
| **North-South (NS)** | 39 |
| **South-North (SN)** | 39 |

Five sample PSMs collected from our vision-based approach are presented as examples in Table 3. The attribute of "Device user type" (see row two of Table 2) is "vulnerable road user (VRU)," which represents a pedestrian. The TempID attribute contains the temporary ID of each pedestrian, and the longitude and latitude contain the location information of each pedestrian. We use a value of 201 m, which is the elevation of the Perimeter Road and Avenue of Champions signalized intersection for the elevation attribute. The positional accuracy is 0.54 m, which is calculated from the sample distribution of latitude and longitude considering 1-sigma standard deviation. For our purposes, we define the heading as text data. For example, the pedestrian with TempID 2 is moving to the North-South (NS) direction as provided in the 3rd column of Table 3.

**Table 3: Example of PSMs generated from our vision-based deep learning approach**

| Data Element | PSM Packet 1 | PSM Packet 2 | PSM Packet 3 | PSM Packet 4 | PSM Packet 5 |
|---|---|---|---|---|---|
| Message ID | 18547 | 20352 | 18548 | 20353 | 18549 |
| Device User Type | VRU | VRU | VRU | VRU | VRU |
| Timestamp | 1543609955382 | 1543609955398 | 1543609955518 | 1543609955541 | 1543609955653 |
| Message Count | 1 | 2 | 3 | 4 | 5 |
| TempID | 1 | 2 | 1 | 2 | 1 |
| Latitude | 34.679183 | 34.679183 | 34.679183 | 34.679183 | 34.679183 |
| Longitude | -82.847414 | -82.84771 | -82.847414 | -82.84771 | -82.847414 |
| Elevation (m) | 201 | 201 | 201 | 201 | 201 |
| Positional Accuracy (m) | 0.54 | 0.54 | 0.54 | 0.54 | 0.54 |
| Velocity (m/s) | 0.040266 | 0.478718 | 0.040266 | 0.478718 | 0.029081 |
| Heading | SN | NS | SN | NS | SN |

## 4.3 Evaluation Results

After collecting the field data, we evaluated the accuracy of the generated PSMs following SAE J2945 standard. As we described in Section 3, we evaluated the accuracy of latitude, longitude, and velocity of a pedestrian as these data change every 100 ms. First, in Section 4.3.1, we evaluated the pedestrian detection accuracy to ensure that a pedestrian must be detected if the pedestrian is in a safety-critical zone of the signalized intersection. Later, in sections 4.3.2 and 4.3.3, we evaluated the accuracy of the pedestrian location and velocity estimation from the vision-based approach, respectively.

### *4.3.1 Accuracy of pedestrian detection*

In our experiment, we used the state-of-the-art YOLOv3 object detection model for detecting pedestrians. Although this YOLOv3 deep learning model is applicable for a real-time safety application (Redmon et al., 2018), here we achieved only a detection accuracy of 81%, which was inadequate for any safety-critical pedestrian detection application. Thus, we retrained the YOLOv3 model with the collected video data for all directions (e.g., East-West, West-East, North-South, and South-East). Before retraining, we modify the output layer of the pre-trained YOLOv3 to detect pedestrians only. Of the 1300 images captured, we used 900 for training and 400 for testing the pedestrian detection accuracy of the YOLOv3. We annotated each image manually to generate ground truth data. Each annotated image of the video feed was in standard Pascal Visual Object Class (VOC) format *(Everingham et al., 2015)*. As described in Section 3.1, the pedestrian detection deep learning method gives an output that contains a bounding box for each pedestrian and the pedestrian detection confidence score. However, given that multiple bounding boxes characterize a single pedestrian, we used the non-max suppression algorithm as described in Section 3.2 to remove the unnecessary bounding boxes and improve accuracy. The pedestrian detection accuracy was measured using the Eq. (12):

$$Accuracy = \frac{TP}{TP + FP} \tag{12}$$

where, $TP$ and $FP$ represent the true positive and false positive detection, respectively. We found 98% $TPs$ and 2% $FPs$ among the 225 observations as shown in Table 2.

### *4.3.2 Accuracy of pedestrian localization*

We compared generated PSMs based on our approach with the data collected from the pedestrian DSRC-enabled device. For this comparison, first, we generated the ground truth location data of pedestrians, which was the actual location and velocity of a pedestrian while walking. A GIS map tool was used to discretely create this data by geocoding the longitude and latitude for each pedestrian in the image. We calculated the Root Mean Square Error (RSME) between PSMs generated from the DSRC-enabled pedestrian hand-held device and ground truth data, and PSMs from our vision-based PASS and ground truth data. We then used this information to evaluate the performance of the generated PSMs using the Eq. (13):

$$RMSE = \sqrt{\frac{\sum_{T=1}^{N}(G_i - P_i)^2}{N}} \quad (13)$$

where, $G_i$ is the actual location of a pedestrian as per ground truth data; $P_i$ is the pedestrian location of PSMs generated from our approach or the DSRC-enabled pedestrian hand-held device, and $N$ is the number of observations. The RMSE is shown in Figure 4, which was measured using the latitude and longitude information generated from our vision-based PASS and DSRC-enabled pedestrian hand-held device. We observed that for each pedestrian heading direction, the distance error between the ground truth and DSRC-enabled pedestrian handheld device is higher than the distance error between the ground truth and pedestrian location information based on our approach. Such inaccuracy can occur from GPS location errors from the device *(Cohda, 2017)*. We found that the location generated from our approach is close to the actual pedestrian location.

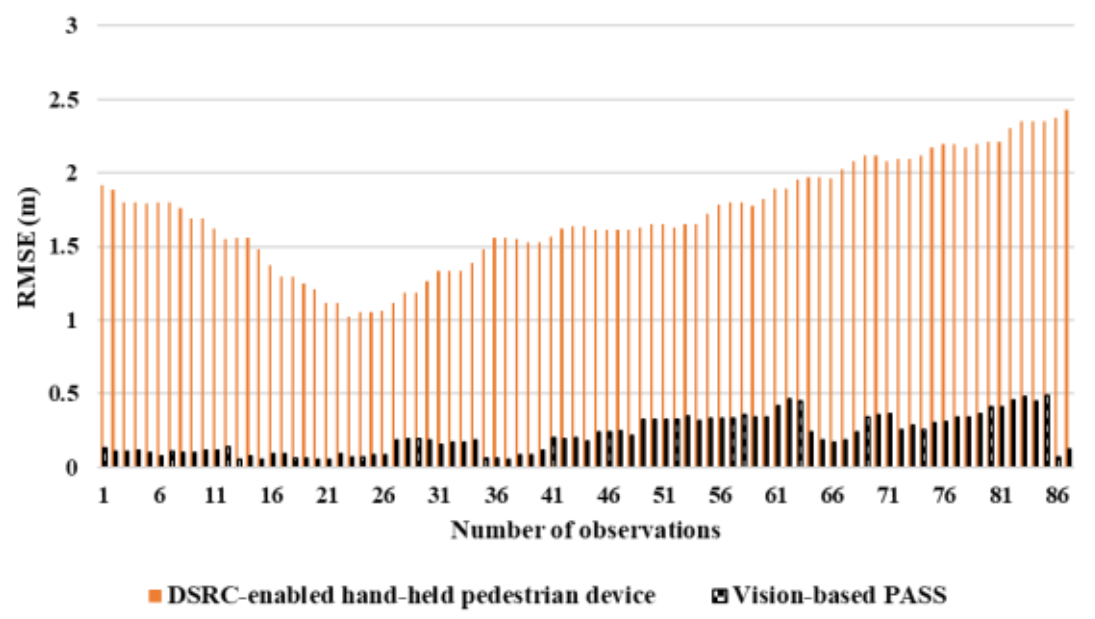

(a) Pedestrian heading direction: East-West

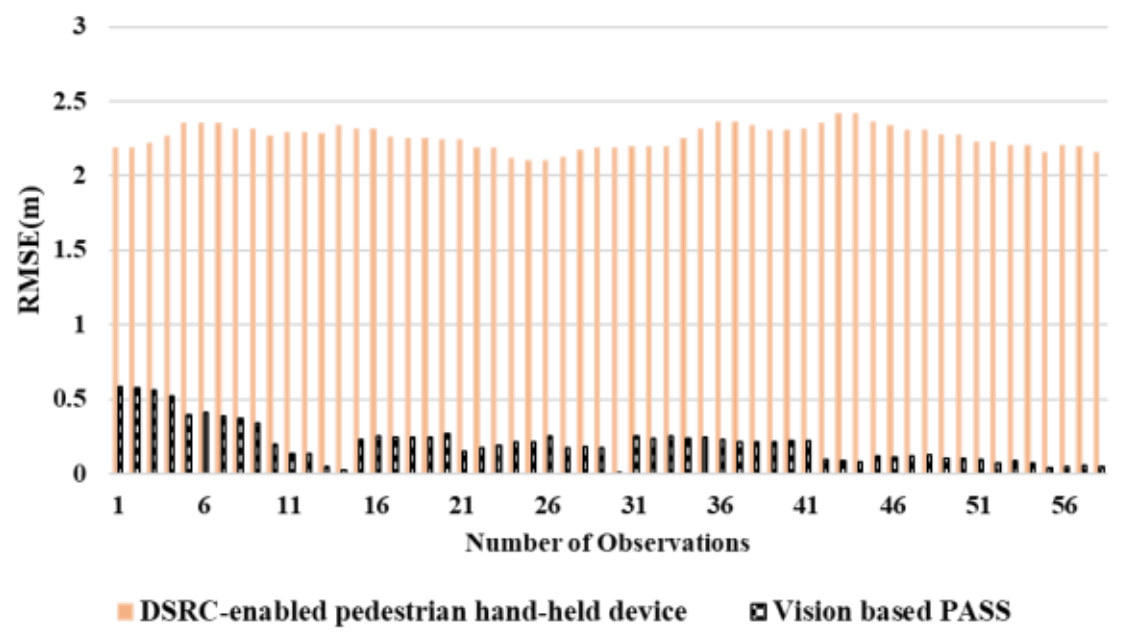

(b) Pedestrian heading direction: West-East

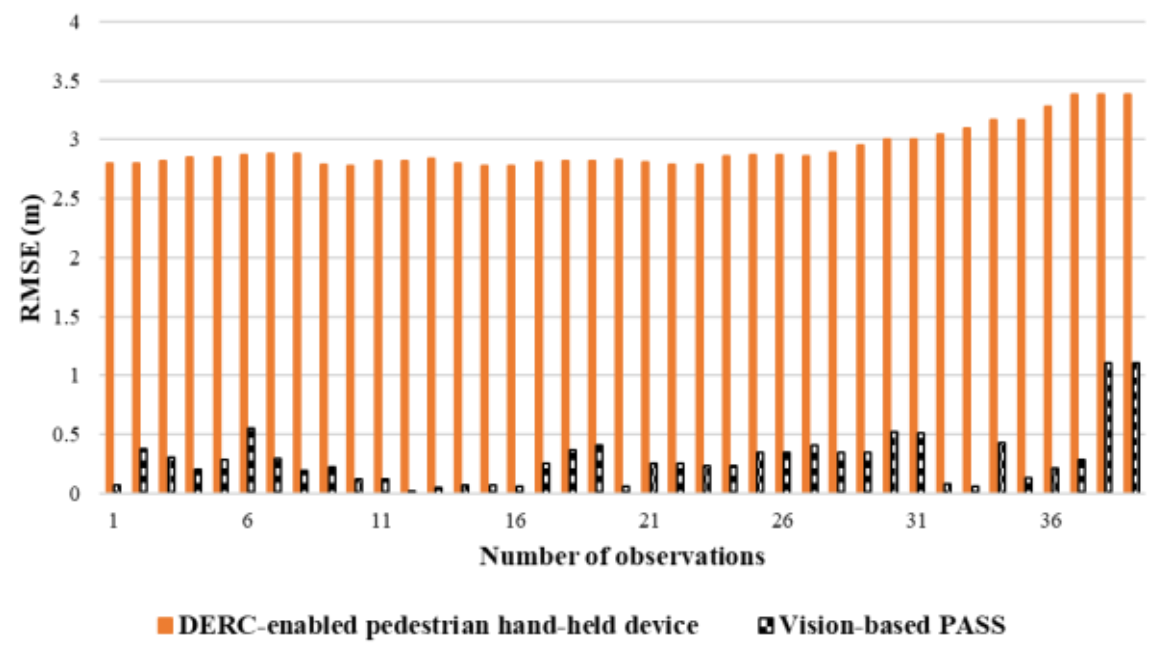

(c) Pedestrian heading direction: North-South

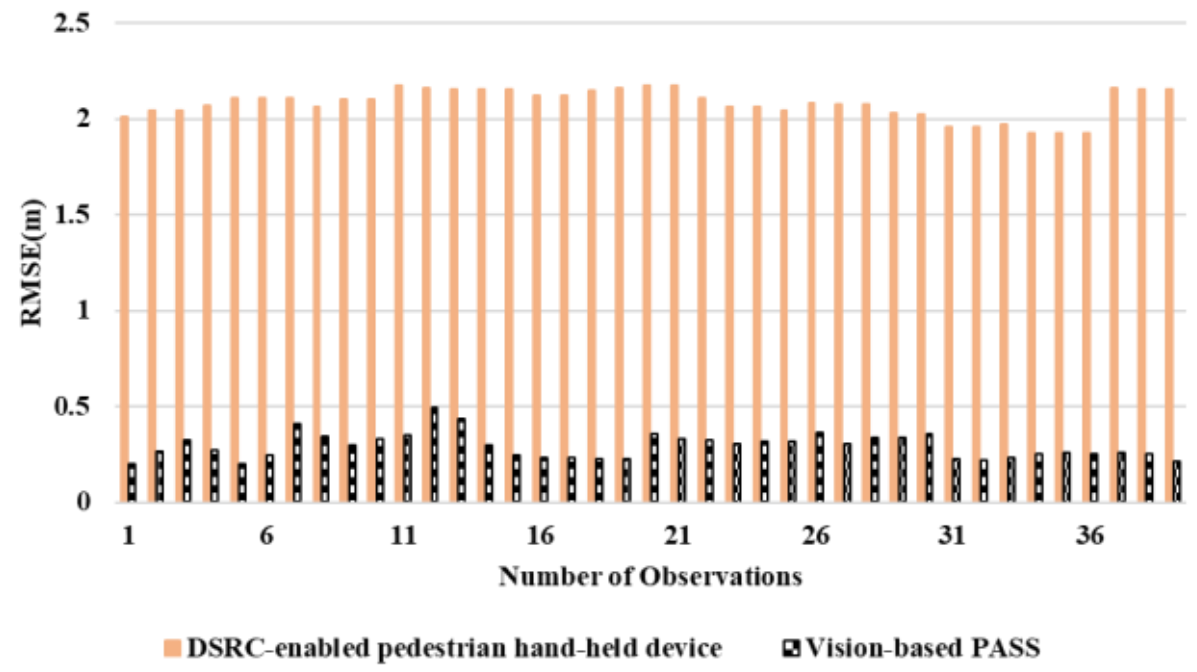

(d) Pedestrian heading direction: South-North

**Figure 4: Location RMSE between our vision-based PASS and DSRC-enabled pedestrian hand-held device compared to the actual pedestrian location.**

We also calculated the average RMSE for DSRC-enabled pedestrian hand-held device and vision-based PASS compared to the actual pedestrian location. The average RMSE for each pedestrian direction is shown in Figure 5. Here, we also observed that the average range of the RMSE using vision-based PASS falls within 0.21 m to 0.31 m, whereas the RMSE using DSRC-enabled hand-held device falls within 1.70 m to 2.25 m. Considering all directions, the positional RMSE is 0.25 m. The maximum acceptable limit for the deviation from the ground truth position value is 1.50 m, which represents the 1-sigma standard deviation, according to the SAE J2945 standard. Thus, the vision-based PASS fulfills the positional accuracy requirement. Furthermore, our analysis indicates the efficacy of the vision-based PASS in locating the pedestrian more accurately compared to the commercially available DSRC-enabled hand-held devices.

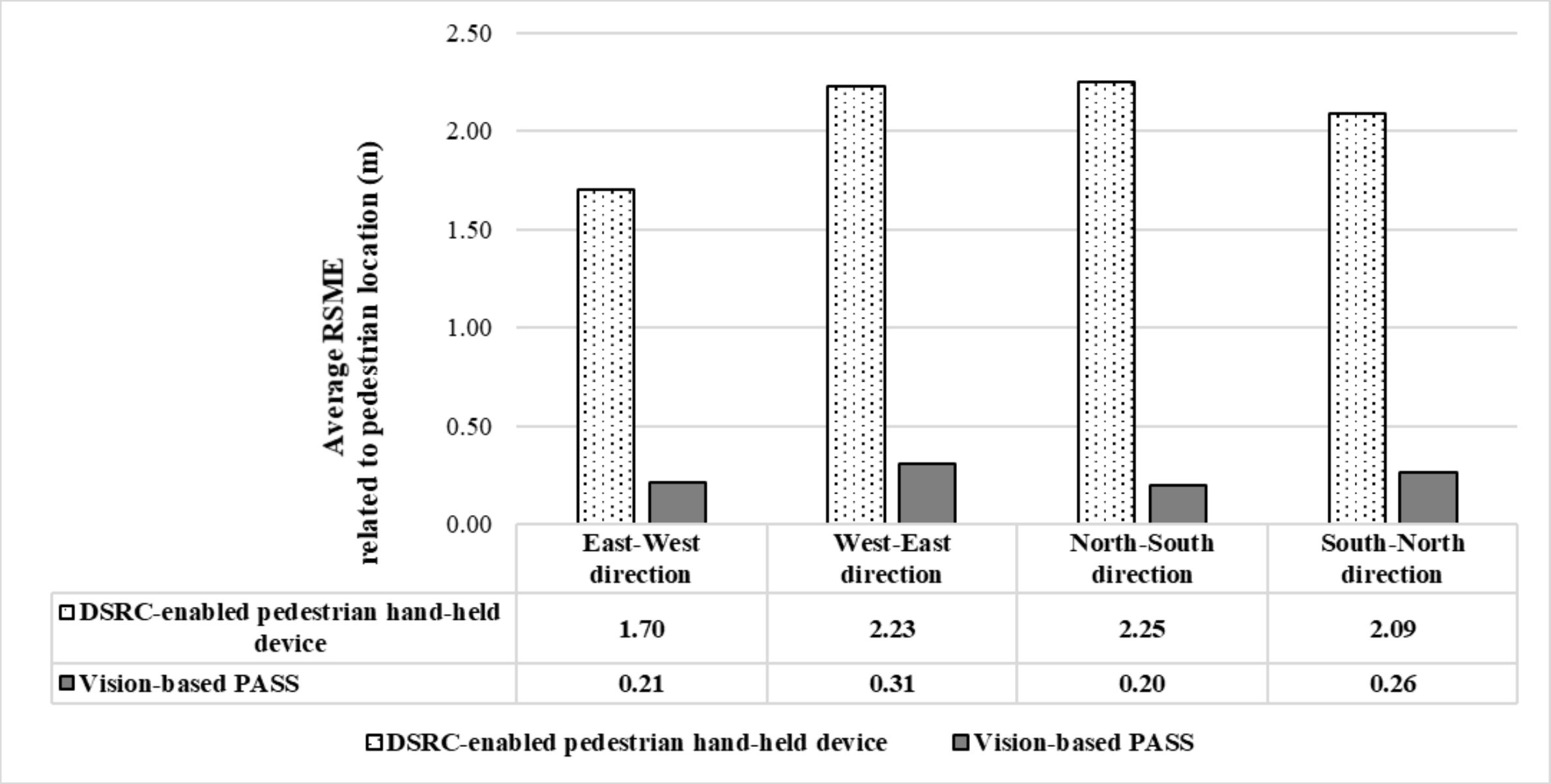

**Figure 5: Average location RMSE between our vision-based PASS and DSRC-enabled pedestrian device compared to the actual pedestrian location.**

### *4.3.3 Accuracy of estimation of the pedestrian velocity*

To evaluate the performance of pedestrian velocity estimation, we use the ground truth velocity as described in Section 4.3.2. The ground truth velocity contains the velocity of each pedestrian in each image. We then compared the RMSE of DSRC-enabled pedestrian hand-held device from the actual ground truth velocity to the RMSE between the vision-based PASS and the actual ground truth velocity of the pedestrian. To measure the velocity accuracy of the RMSE, we used the Eq. (14):

$$RMSE = \sqrt{\frac{\sum_{i=1}^{N}\left(V_{gt\ (t,i)} - V_{t,i}\right)^2}{N}} \tag{14}$$

where, $V_{gt\ (t,i)}$ is the ground truth velocity of the $i^{th}$ pedestrian at time $t$ at any direction; and $V_{t,i}$ is the velocity of the $i^{th}$ pedestrian from the DSRC-enabled hand-held device or from our vision-based PASS at time $t$. The average RMSE for each direction is shown in Figure 6. Here, our vision-based PASS is used to calculate the RMSE of the pedestrian velocity compared to ground truth values of the velocity. We also observed that the average RMSE using our vision-based PASS ranges from 0.33 m/s to 0.46 m/s, whereas the RMSE calculated using DSRC-enabled pedestrian hand-held device ranges from 0.74 m/s to 0.81 m/s. Considering all directions, the average estimated RMSE for velocity is 0.39 m/s for vision-based PASS. A subsequent Z-test analysis with a 95% confidence interval revealed no significant difference of the velocity data based on our approach from the ground truth velocity, thus validating that the vision-based PASS can generate pedestrian velocity accurately with the established interval of 100 ms.

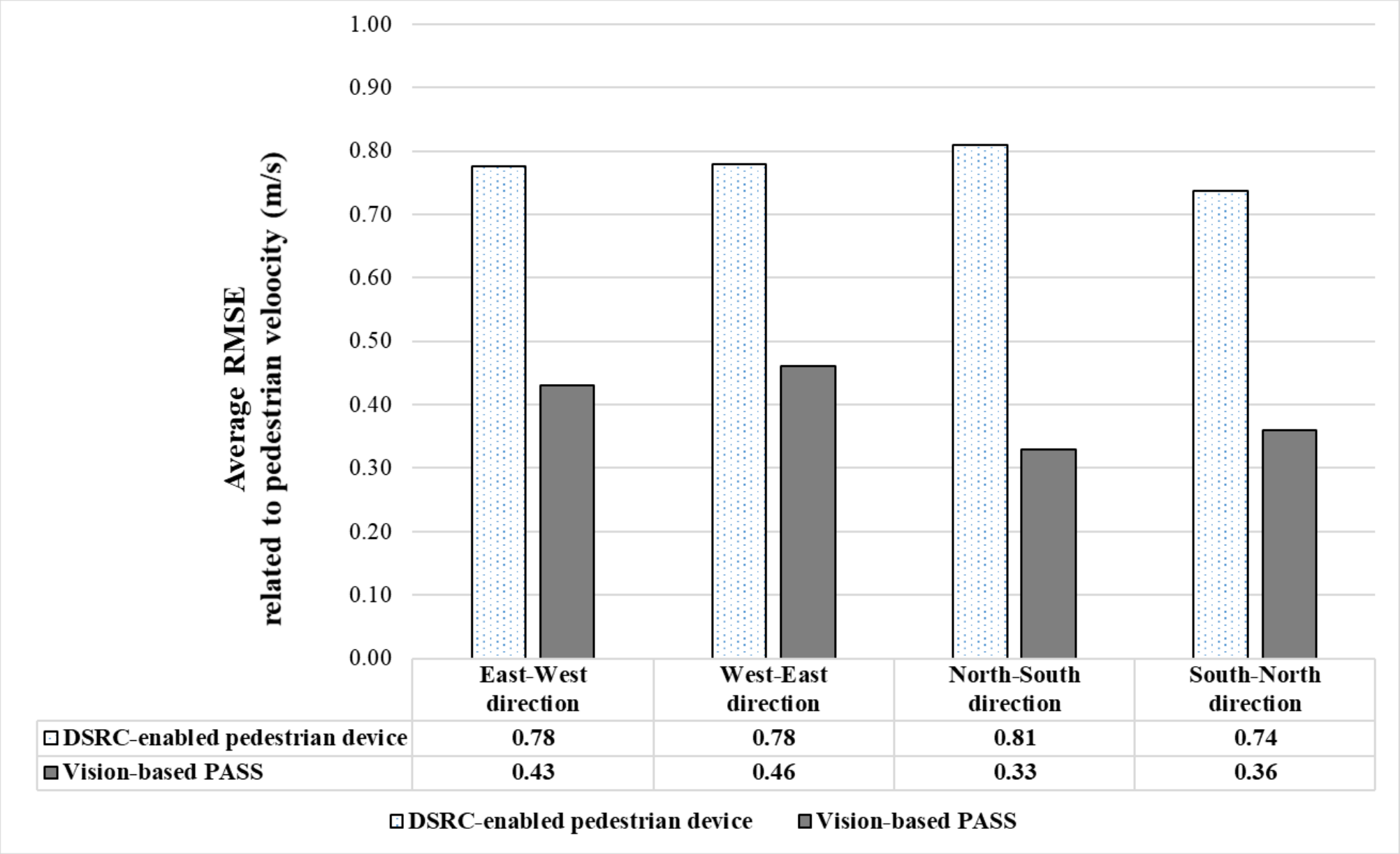

**Figure 6: Velocity RMSE between our vision-based PASS and DSRC-enabled pedestrian device compared to the actual pedestrian velocity.**

## 5. EVALUATION OF PEDESTRIAN SAFETY

In this section, we evaluated our vision-based pedestrian safety with the PSCW application by analyzing pedestrian collision warning or safety alerts. A system level evaluation is necessary for providing a system level justification by measuring the computational latency and end-to-end communication latency. The PSCW application provides a pedestrian safety alert to the approaching connected vehicles from the roadside infrastructure to the connected vehicle so that the vehicle can avoid possible collisions between the vehicle and pedestrian. According to the SAE J2945 standard, a PSCW application output must be delivered to the connected devices with no more than 100 ms *(Karagiannis et al., 2011)*. Thus, possible crash warnings from the PSCW application using generated PSMs by the vision-based approach must deliver to the connected devices and connected vehicles within 100 ms.

### 5.1 Evaluation scenario

In this evaluation scenario, a pedestrian is crossing the road while a connected vehicle is approaching the intersection. Using our vision-based approach, the PSCW application generates a collision warning message using PSMs based on the detected pedestrian and vehicle trajectory calculated from BSMs of a connected vehicle. To generate a collision warning message, we defined a risk zone where the pedestrian and vehicle have a high chance of a disastrous encounter. We defined three critical locations A, B, and C, as shown in Figure 7, to evaluate the PSCW application. Location A indicates a position of a collision between pedestrian and vehicle if the pedestrian continues walking and the vehicle continues without noticing the pedestrian. Location

B indicates a position where the vehicle receives the first collision warning message. Location C indicates a position where the vehicle halts if it receives the first warning message at point B and decelerates from that point. Based on this scenario, we evaluated the efficacy of PSCW application by generating pedestrian collision warnings to avoid potential vehicle-pedestrian collisions. We subsequently evaluate the communication latency to send the pedestrian collision warning or safety alert from the roadside infrastructure to a connected vehicle.

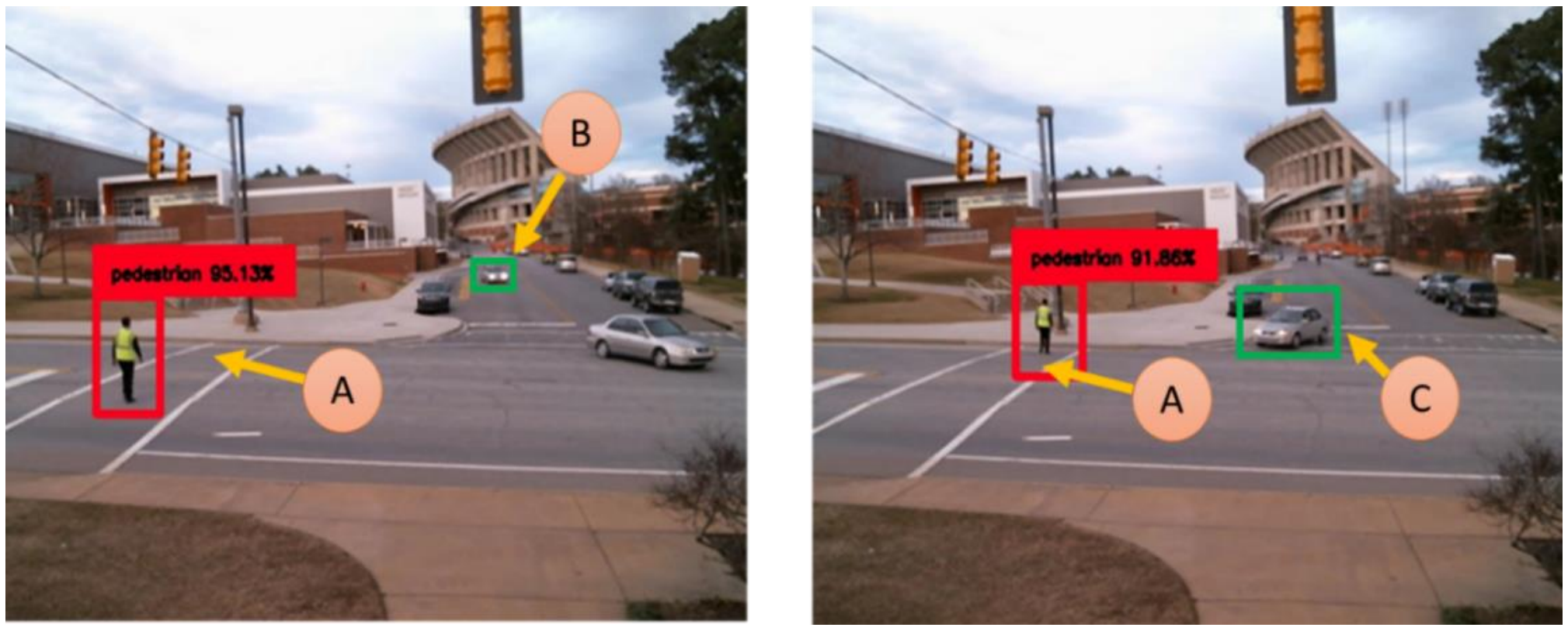

(a) Experimental set-up of the field test scenario for pedestrian collision warnings

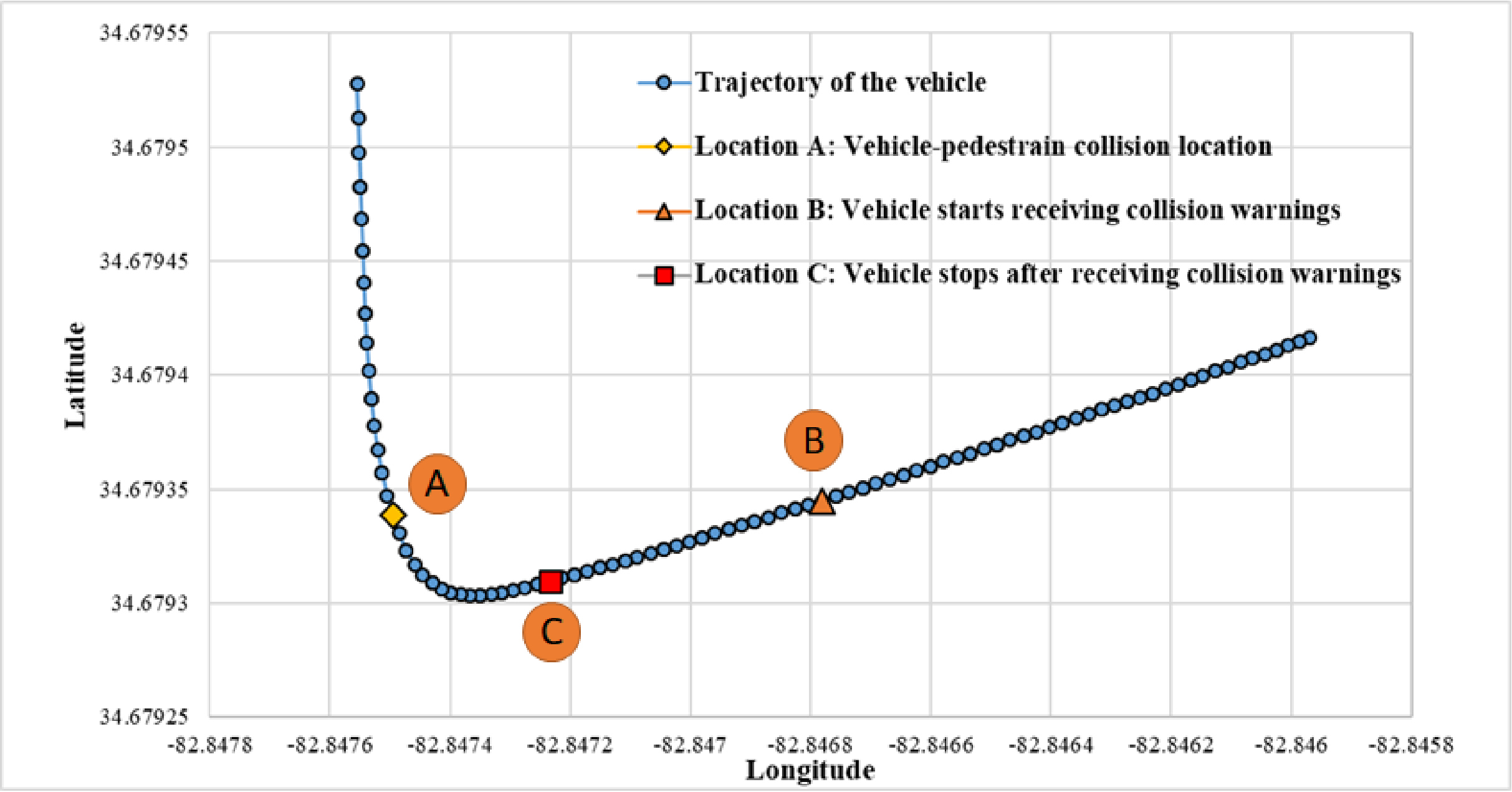

(b) Vehicle trajectory plot of the field test scenario for pedestrian collision warnings

**Figure 7: Evaluation scenario for pedestrian safety using pedestrian collision warnings**

### 5.2 Evaluation Results

We used Time-to-Collision (TTC) matric to evaluate the PSCW application. We defined TTC as the time required for a vehicle to collide with a pedestrian if both the pedestrian and vehicle continue on their present trajectories (e.g., velocity and direction) without any change in their trajectory *(Karamouzas et al., 2009)*. Based on our field collected data, TTC is the time elapsed between the pedestrian enters into the risk zone and reaches the collision point. Our calculated TTC is 7.3 s for the evaluation scenario, as shown in Figure 8. At the starting point of an assumption of T=0 s, the vehicle was located at Point B. Thus, if the vehicle continues to follow its trajectory from point B without noticing the pedestrian, it will collide with the pedestrian, who is coming from point A, after T=7.3 s. However, if the vehicle receives a collision warning notification at point B, and decelerates at 3.35 $m/s^2$, it will travel 46 m before making a full stop (i.e., velocity 0 km/h). Thus, it will reach at point C when the vehicle completely stops, and the vehicle thus avoids the collision. The results from this study prove the efficacy of the pedestrian safety alert system in generating real-time collision warnings to avoid a possible vehicle-pedestrian collision.

### 5.3 Latency Requirement

To evaluate the systems level performance, we measured the end-to-end latency from the time at which the image was captured by the camera to that when a DSRC-enabled connected vehicle received a pedestrian safety alert. Thus, latency consists of computational and communication latency. Computational latency depends on the hardware configuration of the system. For our experimental setup, we used a computer with Intel i7 processor with a 6GB GPU memory to run the pedestrian detection and performs the calculations necessary for generating pedestrian collision warnings. The communication network latency is the time difference between sending a collision warning from one connected roadside device and receiving the same collision warning to another connected device or vehicle. The end-to-end latency distribution is shown in Figure 8, which combines computational latency and communication latency. Also presented are the latency data ranges from 58 ms to 64 ms, which encompasses 75% of the sample data with less than 1% of the pedestrian collision warnings exhibiting end-to-end latency above 80 ms.

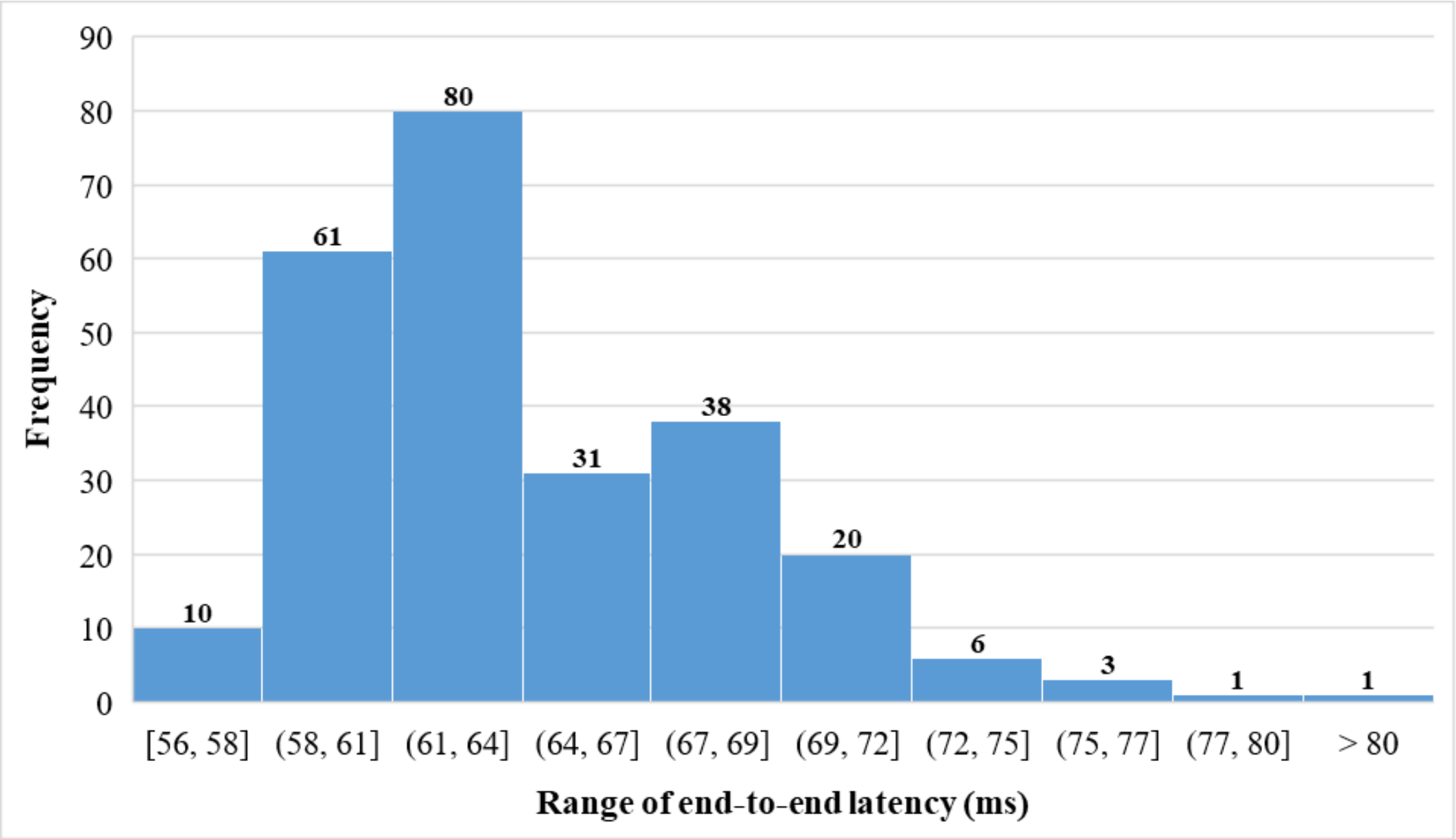

**Figure 8: Distribution of end-to-end latency for the pedestrian collision warnings**

**Table 4: Computational and communication network latency of vision-based PASS**

| Latency Type | Minimum Latency (ms) | Maximum Latency (ms) | Average Latency (ms) | Maximum Allowable Latency (ms) |
|---|---|---|---|---|
| **Computational Latency** | 54 | 64 | 56 | 100 ms |
| **Communication Network Latency** | 2 | 17 | 4 | |

The computational latency and communication network latency are summarized in Table 4. Note that an average latency of the DSRC-enabled hand-held devices is only 4 ms, which includes the communication network latency. Our field experiments show that the vision-based PASS is characterized by an average latency of 60 ms as it includes a computational latency atop the communication network latency. The total maximum latency that combines both the computational and communication latency is below the required 100 ms requirement of pedestrian safety alert dissemination from the RSU to the DSRC-enabled connected devices. Thus, our vision-based PASS fulfills the latency requirement for safety-critical applications *(Karagiannis et al., 2011).*

## 6. CONCLUSIONS

Our research is focused on the creation of a novel real-time vision-based approach to improve pedestrian safety through the accurate detection of pedestrians, the generation of personal safety messages (PSMs) and providing pedestrian safety alerts. Our vision-based pedestrian alert safety system (PASS) can generate personal safety messages (PSMs) and safety alerts in real-time (every 100 milliseconds) using generated PSMs to improve pedestrian safety at a signalized intersection. Analyses results revealed that our vision-based PASS can estimate the location and velocity of a pedestrian more accurately in terms of RMSE compared to existing DSRC-enabled hand-held pedestrian devices. Furthermore, we evaluate the vision-based pedestrian safety at a system level by conducting a real-world field experiment using a connected vehicle PSCW application The system-level evaluation of our PASS demonstrates that pedestrian detection through the analysis of video data generates accurate PSMs and safety alerts in real-time. Numerical analyses from our experiment show that our vision-based PASS improves pedestrian safety. However, to generate accurate pedestrian safety alerts, it is required to train the vision-based deep learning model with different weather and lighting conditions for each signalized intersection.

## ACKNOWLEDGMENT

This material is based on a study supported by a grant from the Center for Connected Multimodal Mobility ($C^2M^2$) (USDOT Tier 1 University Transportation Center) headquartered at Clemson University, Clemson, South Carolina, USA. Any opinions, findings, and conclusions or recommendations expressed in this material are those of the author(s) and do not necessarily reflect the views of the Center for Connected Multimodal Mobility ($C^2M^2$), and the U.S. Government assumes no liability for the contents or use thereof.